\documentclass{article}
\usepackage{spconf,amsmath,graphicx,amssymb,tikz}

\newcommand\copyrighttext{%
  \footnotesize \copyright 2018 IEEE. Personal use of this material is permitted. Permission from IEEE must be
obtained for all other uses, in any current or future media, including
reprinting/republishing this material for advertising or promotional purposes, creating new
collective works, for resale or redistribution to servers or lists, or reuse of any copyrighted
component of this work in other works.}
\newcommand\copyrightnoticeb{%
\begin{tikzpicture}[remember picture,overlay]
\node[anchor=south,yshift=10pt] at (current page.south) {\fbox{\parbox{\dimexpr\textwidth-\fboxsep-\fboxrule\relax}{\copyrighttext}}};
\end{tikzpicture}%
}

\newcommand{\R}{\mathbb{R}}                      
\newcommand{\Omit}[1]{}						     

\title{Endmember Extraction on the Grassmannian}
\name{Elin Farnell, Henry Kvinge, Michael Kirby, Chris Peterson
\thanks{This paper is based on research partially supported by the National Science Foundation under Grants No. DMS-1513633, and DMS-1322508.}}
\address{Colorado State University\\Department of Mathematics\\Fort Collins, CO 80523-1874}
\begin{document}
%
\maketitle

\copyrightnoticeb
\begin{abstract}
Endmember extraction plays a prominent role in a variety of data analysis problems as endmembers often correspond to data representing the purest or best representative of some feature. Identifying endmembers then can be useful for further identification and classification tasks. In settings with high-dimensional data, such as hyperspectral imagery, it can be useful to consider endmembers that are subspaces as they are capable of capturing a wider range of variations of a signature. The endmember extraction problem in this setting thus translates to finding the vertices of the convex hull of a set of points on a Grassmannian. In the presence of noise, it can be less clear whether a point should be considered a vertex. In this paper, we propose an algorithm to extract endmembers on a Grassmannian, identify subspaces of interest that lie near the boundary of a convex hull, and demonstrate the use of the algorithm on a synthetic example and on the 220 spectral band AVIRIS Indian Pines hyperspectral image.
%
\end{abstract}
\section{Introduction}
\label{sec:intro}

There is increasing evidence that for certain types of high-dimensional data, particularly when the data contains samples of similar objects in a variety of state, it is advantageous to work with subspaces of the ambient data space rather than individual points. Examples of this include improved classification of signals in hyperspectral images \cite{CK14,chepushtanova2017sparse} and identification of ultra-low resolution grayscale face images via illumination spaces \cite{CKKPDB07}. In these examples, the underlying geometric framework is that of Grassmann manifolds. The real Grassmann manifold $Gr(k,n)$ is a topological manifold whose points parameterize all possible $k$-dimensional subspaces of $\mathbb R^n$. There are many distinct ways to impose a geometric structure on $Gr(k,n)$. The most commonly used include geometries imposed by the geodesic, chordal, and Fubini-Study metrics. Each of these geometries has its own distinct characteristics and its own distinct advantages and disadvantages. Although the geometry of the Grassmannian, with respect to each of these various metrics, is well understood from a theoretical perspective, many of the standard methods for studying data in Euclidean space do not yet have analogues on the Grassmannian. In this paper we will describe a method for computing endmembers and other points of interest within a collection of points on the Grassmann manifold $Gr(k,n)$. 

Given a set of data points $X$ in $\R^n$, one can consider the convex hull $C$ of $X,$ which is defined as the smallest convex set containing $X$. If $X$ is finite then $C$ will be a convex polytope and one can ask for the vertices $x_1, x_2, \dots, x_\ell$ of this polytope. The vertices $x_1, x_2, \dots, x_\ell$ are sometimes known as {\emph{endmembers}}. Interest in endmembers stems from their relationship to extrema of linear functionals on X and thus endmember identification exposes extremal elements in $X$. Note that for any datapoint $y$ lying in the convex hull of $X$ we can express $y$ as a convex combination of its endmembers, i.e. there exists an expression
\begin{equation} \label{eqn-linear-mixing-model}
y = \sum_{i=1}^\ell w_i x_i
\end{equation}
where $w_i\geq 0$ and $w_1 + \dots + w_\ell = 1$. Thus, in some sense, the endmembers of $X$ represent the most novel signals in $X$ and every other point in $X$ can be recognized as a convex combination of these fundamental elements. The coefficients $w_1, w_2, \dots, w_\ell$ are known as the fractional abundances of $x_1, \dots, x_\ell$ in $y$. The determination of endmembers of a hyperspectral image is a particularly well-studied problem which is a key step of the more general problem of decomposing each pixel into its pure signals, also known as the spectral unmixing problem \cite{BPDPDGC12}. There are a significant number of algorithms for endmember extraction. These include the pixel purity index (PPI) \cite{BKG95}, N-FINDR \cite{Win99},  optical real-time adaptive spectral identification system (ORASIS) \cite{BPABR95}, iterative error analysis (IEA) \cite{neville1999automatic}, convex cone analysis (CCA) \cite{IC99}, vertex component analysis \cite{ND05}, orthogonal subspace projection (OSP) technique \cite{HC94}, automated morphological endmember extraction (AMEE) \cite{PMPP02}, and simulated annealing algorithm (SAA) \cite{BAW00}. We note in particular that the algorithm presented in this paper can be seen to address some of the same challenges that the SAA algorithm recognizes; namely those of endmember variability.

While the algorithm described in this paper might be accurately described as non-linear endmember extraction, the framework is different from other established methods falling under this designation \cite{DTRBMH14}. Motivating the latter is the observation that under certain circumstances, the spectral interactions between endmembers in hyperspectral images can be non-linear. In other words, the data is sometimes better modeled as living on a non-linear submanifold (unknown before-hand) which is embedded in $\R^n$. In our case, the non-linearity arises from the fact that we take subspaces of $\R^n$ rather than single points. We consider these as points on a Grassmannian, thus giving rise to an immediate non-linear framework (irrespective of the particularities of the data).

\section{Endmember Extraction Algorithm}
\label{sec:Algorithm}

We design our endmember extraction algorithm around the exploitation of an isometric embedding of a Grassmannian manifold into Euclidean space followed by a projection into a lower dimensional Euclidean space. Once in this setting, we can utilize known algorithms for discovering vertices of a convex hull. Consider a set of points $X=\left\{x_i\right\}_{i\in \mathcal{I}}$ on the Grassmannian $Gr(k,n).$ We define vertices of the convex hull of $X$ on $Gr(k,n)$ to be those points in $X$ with indices $\mathcal{I}'\subset\mathcal{I}$ obtained in the following way.
\begin{enumerate}
\item Construct a distance matrix $D$ for $X$ using chordal distance on $Gr(k,n)$ (the projection Frobenius norm).
\item Use Multidimensional Scaling (MDS) to find an embedding $E$ for $D$ that is an isometry.
\item Apply the Convex Hull Stratification Algorithm (CHSA) to identify the indices $\mathcal{I}'$ of the vertices of the convex hull in Euclidean space.
\item Via the bijection between $X$ and $E,$ the indices $\mathcal{I}'$ are precisely the indices for points in $X$ that are vertices of the convex hull of $X$ on $Gr(k,n).$
\end{enumerate}

To be precise, we elaborate on these steps here. Chordal distance is defined as follows. If $x_i,x_j\in Gr(k,n),$ then $x_i, x_j$ correspond to two $k$-dimensional subspaces of $\mathbb R^n$. The chordal distance $d$ is defined as $$d(x_i,x_j)=\sqrt{\sum_{m=1}^k\sin^2\theta_m},$$ where $\theta_1,\ldots,\theta_k$ are the principal angles between the subspaces $x_i$ and $x_j.$ For convenience, one may compute chordal distance in the following way. If $S,T$ are orthonormal bases for two $k$-dimensional subspaces $x_i$ and $x_j$ in $\mathbb{R}^n,$ then $$d(x_i,x_j)=\sqrt{k-\|S^T T\|_F^2}\hspace{1mm},$$ where $\|\cdot\|_F$ is the Frobenius matrix norm. For more context on the chordal distance metric, see, e.g., \cite{CHS96,DHST08}. The choice of chordal distance as a metric on the Grassmannian is notable; as shown in \cite{CHS96}, the Grassmannian can be isometrically embedded when this metric is used.

Multidimensional scaling provides a means of defining a collection of points in Euclidean space whose pairwise distances are as faithful as possible to their original distances on some manifold. See \cite{M78,T52,T58,G66,S35,Y38} for full details. The procedure is described briefly below.
Given a $p\times p$ distance matrix $D,$ compute matrices $A$ and $B,$ where $A_{ij}=-\frac{1}{2}D_{ij}^2,$ and $B$ is the result of double-centering $A.$ That is, if $H=I_p-\frac{1}{p}\mathbf{1}\mathbf{1}^T,$ then $B=HAH.$ The MDS embedding is provided via the eigenvectors and eigenvalues of $B.$ Specifically, if $V\Lambda V^{-1}$ is the eigendecomposition of $B,$ then the rows of the matrix $E=\left[\sqrt{\lambda_1}\mathbf{v}_1 \sqrt{\lambda_2}\mathbf{v}_2 \cdots \sqrt{\lambda_q}\mathbf{v}_q \right]$ give a collection of $p$ points in $\mathbb{R}^q$ whose pairwise distances come as close as possible to $D$ provided each $\lambda_1,\ldots,\lambda_q>0.$ 

The Convex Hull Stratification Algorithm proposed in \cite{ZKP2017} and \cite{Z13} utilizes an optimization problem that attempts to represent each point in a data set as a convex combination of its $N$ nearest neighbors. Note that vertices of the convex hull cannot be represented as a convex combination of its $N$ nearest neighbors for any $N$. Those points whose coefficient vectors require at least one negative component are deemed to be points of interest and include the vertices of the convex hull. Further, the norm of the coefficient vector can be used to stratify the points by their proximity to the boundary of the convex hull. Specifically, one solves the following optimization problem for each $\mathbf{x}$ in the data set: $$\min_{\mathbf{w}}\gamma\|\mathbf{w}\|_2^2+\lambda\|\mathbf{w}\|_1+\left\|\mathbf{x}-\sum_{j\in N}w_j\mathbf{x}_j\right\|_2^2$$ subject to $\sum_{j\in N} w_j=1,$ where the set $\left\{\mathbf{x}_j\right\}_{j\in N}$ is the set of $N$ nearest neighbors to $\mathbf{x}$ in the data set. Those vectors $\mathbf{x}$ with an associated vector $\mathbf{w}$ for which at least one component of $\mathbf{w}$ is less than zero necessarily contain the vertices of the convex hull. Additionally, a measure of boundary-proximity is provided by the $\ell_2$-norm of the associated vectors $\mathbf{w}.$

To summarize, our algorithm exploits an isometric embedding of the Grassmannian into Euclidean space, where well-known convex hull algorithms may be applied. Thus it is computationally manageable and serves as a useful technique in applied settings where a ``pure representative" for a feature may be better represented as a span of several realizations of the ``pure representative."

The computational complexity of the Grassmann Endmember Extraction Algorithm is dominated by the complexity of MDS and CHSA as well as that of constructing a distance matrix for points on $Gr(k,n).$ Classical MDS is known to be $O(n^3)$ but faster versions exist (see, e.g. \cite{C96}). For details on the complexity of CHSA, see \cite{ZKP2017}. A chordal distance matrix can be computed in polynomial time. Combining these three complexities, we conclude that the Grassmann Endmember Extraction Algorithm is a polynomial time algorithm.

\section{Application: Simplex Embedding}
\label{sec:ToyExperiment}

Given a set of subspaces $A_1, A_2, \dots, A_r$ of $\mathbb R^n$, a Singular Value Decomposition (SVD) based algorithm can be used to compute a weighted flag mean of the set of $A_i$ \cite{draper2014flag}. The underlying theory for the construction of the flag mean is based on the chordal metric as this allows for the very direct (and fast) SVD-based algorithm. The input for the algorithm consists of orthonormal bases for each of $A_1, A_2, \dots, A_r$ together with real-valued weights $a_1, a_2, \dots, a_r$. The output of the flag mean algorithm consists of an ordered sequence $u_1, u_2, \dots, u_l$ of orthonormal vectors (defined only up to sign where $l$ is the dimension of the span of all of the $A_i$ considered together). From the $u_i$ one can construct a full flag of nested vector spaces $V_1\subset V_2\subset \dots \subset V_l$ by defining $V_i=span\{u_1, u_2, \dots, u_i\}$. 

If we restrict $A_1, A_2, \dots , A_r$ to all have the same dimension $k$ then for each choice of weights we can consider the $k$-dimensional component of the flag $V_k$. In this manner, we can think of $V_k$, as determined by the weighted flag mean, as representing a kind of convex combination of the $A_i$ as points on the Grassmann manifold $Gr(k,n)$.

In the following example, we consider $3$ random points $A_1, A_2, A_3 \in Gr(3,10);$ then we determine 4997 random weighted flag means of these points, for a total of 5000 points. We use Grassmann endmember extraction to recover the three $3$-dimensional subspaces of $\mathbb R^{10}$ that were used to construct the data set. 

\begin{figure}

\begin{minipage}{1.0\linewidth}
  \centering
  \centerline{\includegraphics[width=9cm]{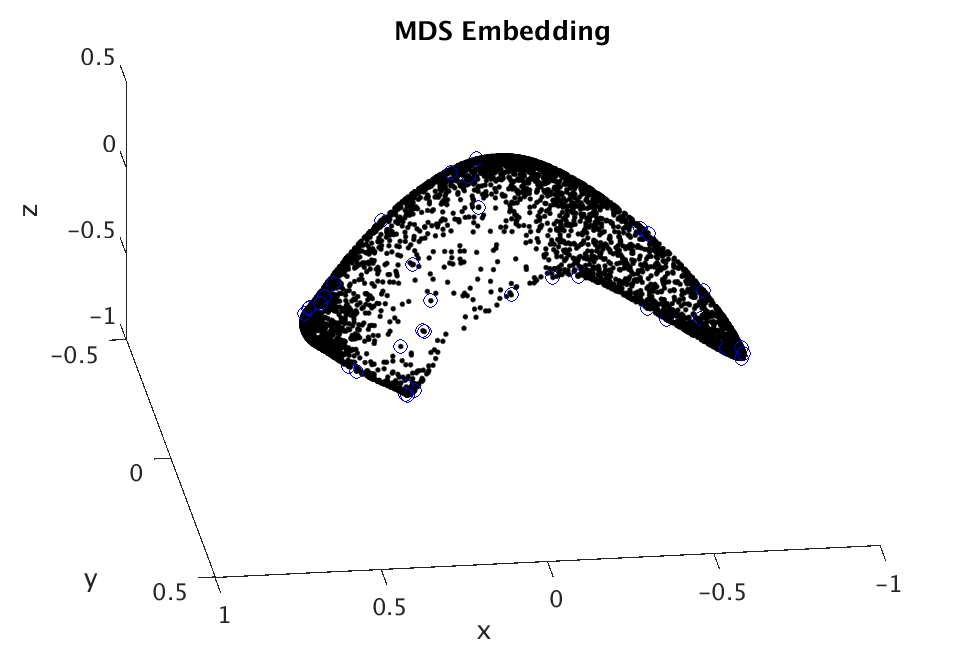}}
\end{minipage}
\caption{MDS Embedding: Simplex. Points with largest $\ell_2$-norm of the coefficient vector are marked with a blue circle. Note that the three vertices on the Grassmannian that were used to generate the data on the manifold are detected as endmembers in the embedding.}
\label{fig:toy2}
\end{figure}

In Figure \ref{fig:toy2}, we show the results of the MDS mapping of the set of points on $Gr(3,10)$ described above to its best $3$-dimensional space. Note that the mapping appears to faithfully represent the geometry and distance relationships present in the original points on the Grassmannian.

After applying our algorithm to this data set, we threshold by the norm of the weight vector and display those points that are thus identified as endmembers on the Grassmannian (Figure \ref{fig:toy2}). The algorithm successfully identified the vertices that generated the data set, providing a proof of concept of the algorithm.


\section{Application: Indian Pines Hyperspectral Imagery}
\label{sec:IndianPines}

We apply our algorithm to the Indian Pines data set and present a visualization of the extracted endmembers in that context.

The Indian Pines data set consists of 220 spectral bands at a resolution of $145\times 145$ pixels. We use a corrected data set that has bands from the region of water absorption removed, resulting in 200 remaining spectral bands.\footnote{The data is available at http://www.ehu.eus/ccwintco/index.php/
Hyperspectral\_Remote\_Sensing\_Scenes.} The data was collected with an Airborne Visible/Infrared Imaging Spectrometer (AVIRIS) at the Indian Pines test site in Indiana \cite{PURR47}. The scene contains various classes, such as woods, grass, corn, alfalfa, and buildings. 

\begin{figure}[htb]

\begin{minipage}[b]{1.0\linewidth}
  \centering
  \centerline{\includegraphics[width=7.0cm]{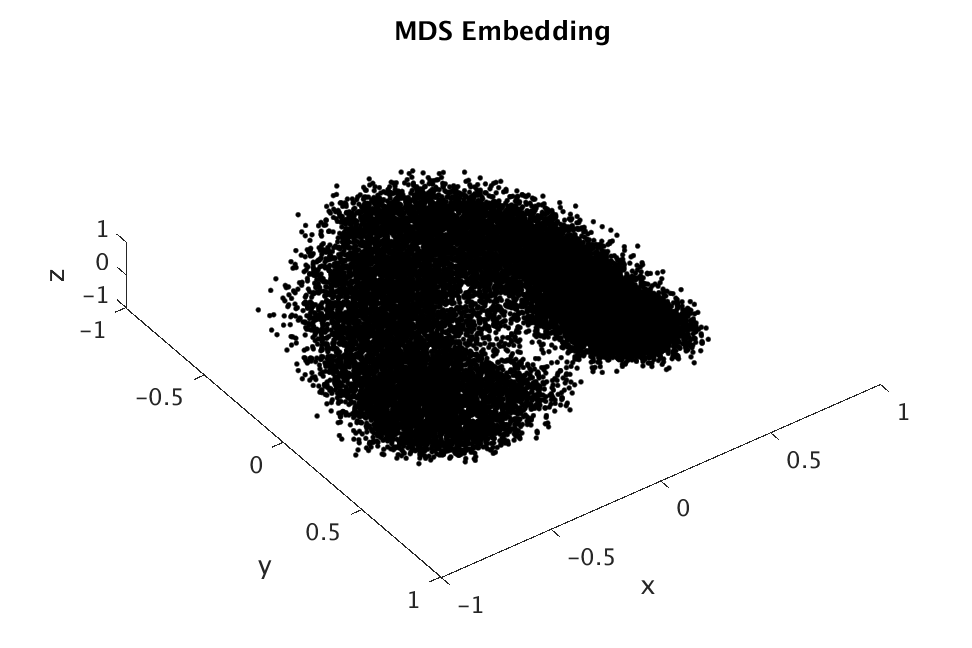}}
\end{minipage}
\caption{Indian Pines Local Patches: MDS Embedding. All subspaces in $Gr(9,200)$ constructed from local $3\times 3$ patches are shown after embedding into $\mathbb{R}^3.$ We apply CHSA to identify vertices using stratification provided by the $\ell_2$-norm of the coefficient vector.}
\label{fig:IPembedding}
\end{figure}

We begin by obtaining points on the Grassmannian by collecting local patches of spectral data. We choose to take regions of size $3\times 3$ pixels. The span of the 9 vectors in a fixed region determine a point in $Gr(9,200).$ Define this set to be $L\subset Gr(9,200).$ We compute pairwise chordal distances between elements of $L$ on $Gr(9,200)$ and define a distance matrix $D.$ The corresponding MDS embedding into $\mathbb{R}^3$ is shown in Figure \ref{fig:IPembedding}. 

\begin{figure}[t!]

\begin{minipage}[b]{1.0\linewidth}
  \centering
  \centerline{\includegraphics[width=7.0cm]{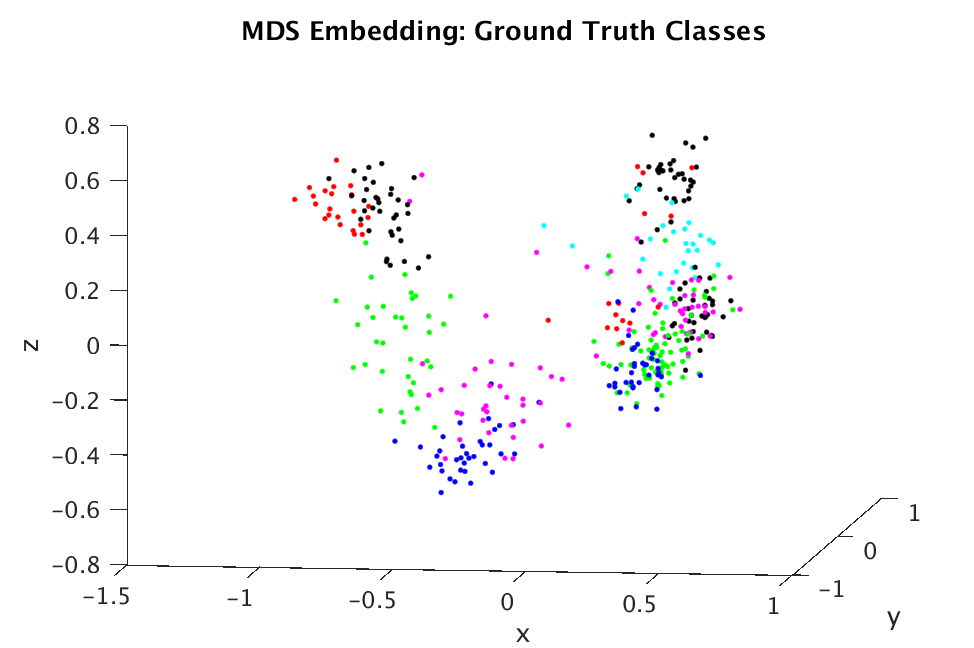}}
\end{minipage}
%
\caption{MDS Embedding: 16 Classes in Indian Pines Scene. The colors correspond to various substances in the scene, such as corn, alfalfa, and buildings. For visibility, colors are used for more than one substance but are chosen to make different classes visually distinguishable. Importantly, this embedding of the points on the Grassmannian demonstrates that appropriate manifold distances and embeddings have the potential to separate classes.}
\label{fig:Classes}
\end{figure}

For comparison, consider the embedding shown in Figure \ref{fig:Classes}. Here, we define points on $Gr(9,200)$ by taking spectral information from random collections of pixel locations, where each draw is made without replacement from a set of pixels that has been manually identified to consist of a particular class (e.g. corn, alfalfa). Call this set $C.$ In Figure \ref{fig:Classes}, an effort has been made to color embedded points by class in such a way that the classes are visually distinguishable. Notably, points within the same class tend to cluster together. Thus, it is reasonable to expect that endmembers extracted from the embedding of spectral information from local patches are likely to correspond to pure forms of various classes.

\begin{figure}[htb]

\begin{minipage}[b]{1.0\linewidth}
  \centering
  \centerline{\includegraphics[width=7.5cm]{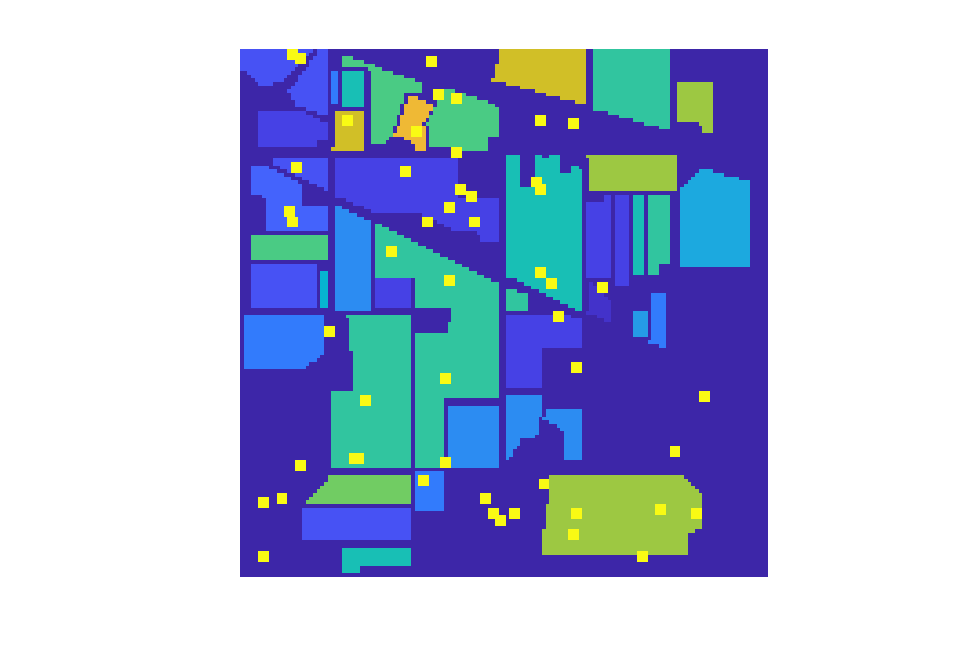}}
\end{minipage}
\caption{Extracted Vertices: Local Patches in Indian Pines Data Set with Largest Coefficient-Vector Norm. These are identified as potential areas of interest for substance purity.}
\label{fig:patches}
\end{figure}

In Figure \ref{fig:IPembedding}, we show the embedding of $L$ into $\mathbb{R}^3.$ We note that by varying parameters in the application of CHSA, one may obtain varying numbers of identified vertices. Thus, depending on the application, one may choose to identify all points that are `near' the boundary or to identify only those vertices that are minimally necessary to define a convex hull. In our setting, we choose to define the number of nearest neighbors to be 7, the $\ell_2$-norm parameter $\gamma$ to be $10^{-10}$ and the $\ell_1$-norm parameter $\lambda$ to be $10^{-5}.$

We can gain some insight into the meaning of the extracted endmembers by visualizing the locations of the local patches that gave rise to those points on the Grassmannian. In Figure \ref{fig:patches}, we show the local patches whose corresponding subspaces are vertices. We conjecture that these locations carry valuable information about various substance classes.

\section{Conclusion}
\label{sec:conclusion}
The convex hull of a set of data points in the manifold setting has significant meaning for a variety of applications. We propose a method that incorporates manifold geometry appropriately in the setting where data naturally lives on a Grassmann manifold. Our method employs an approximately isometric embedding in Euclidean space to identify vertices of the convex hull of a set of points on the Grassmannian. We provide a proof-of-concept of the algorithm in a synthetic example, where we successfully extract the vertices used to generate a convex collection of points on a Grassmannian. We further demonstrate the potential usefulness of the algorithm via an example with hyperspectral imagery, where the identified points on the Grassmannian correspond to meaningful locations in the Indian Pines scene.

\bibliographystyle{IEEEbib.bst}
\bibliography{refs}

\end{document}